\documentclass{article}
\usepackage{amstext}
\usepackage{amssymb}
\usepackage{booktabs}
\usepackage{amsmath}
\usepackage{multirow}
\usepackage{graphicx}
\usepackage{capt-of}
\usepackage[export]{adjustbox}
\usepackage{enumitem}
\usepackage{bbm}
\usepackage{makecell}
\usepackage{threeparttable}
\usepackage{float}

\usepackage[final]{corl_2025} 

\title{Toward Real-World Cooperative and Competitive Soccer with Quadrupedal Robot Teams}

\author{
  Zhi Su$^{1,2,*}$, Yuman Gao$^{1,3,*, \dag}$, Emily Lukas$^{1, *}$, Yunfei Li$^{2}$, Jiaze Cai$^{1}$, Faris Tulbah$^{1}$,\\
  \bfseries Fei Gao$^{3}$, Chao Yu$^{2}$, Zhongyu Li$^{1, \ddag}$, Yi Wu$^{2, 4, \ddag}$, Koushil Sreenath$^{1, \ddag}$\\
  \textsuperscript{1} University of California, Berkeley \quad
  \textsuperscript{2} Tsinghua University \quad
  \textsuperscript{3} Zhejiang University\\
  \textsuperscript{4} Shanghai Qi Zhi Institute \quad
  $^{*}$Equal Contributions. \quad $^{\ddag}$Equal Advising. \\$^{\dag}$Corresponding Author {\tt\small ymgao@zju.edu.cn}. 
}

\begin{document}
\maketitle


\begin{figure}[h]
	\vspace{-0.9cm}
	\centering
	\begin{center}
		\includegraphics[width=1.0\columnwidth]{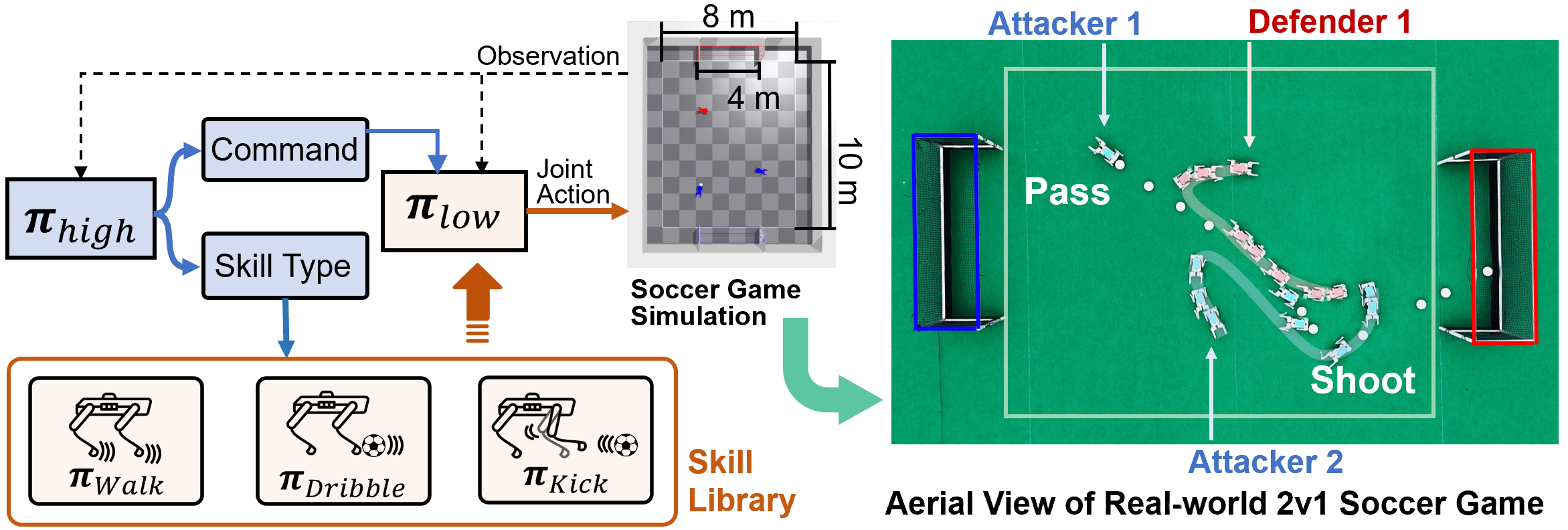}
	\end{center}
	\vspace{-0.0cm}
    \caption{The proposed hierarchical framework consists of a high-level strategy policy ($\boldsymbol\pi_{\text{high}}$) that selects low-level skills and issues corresponding commands, and low-level skill policies ($\boldsymbol\pi_{\text{low}}$) that execute motor primitives including walking, dribbling, and kicking. Policies are trained in simulation and deployed on real quadrupedal robots. In the figure on the right, we show a case of a real-world 2v1 soccer game, where Attacker 1 passes to Attacker 2, who then shoots and scores. For more experimental results, see \url{ https://youtu.be/7gq7N16jKgI }.}
	\label{fig:teaser}
	\vspace{-0.0cm}
\end{figure}

\begin{abstract}

Achieving coordinated teamwork among legged robots requires both fine-grained locomotion control and long-horizon strategic decision-making. Robot soccer offers a compelling testbed for this challenge, combining dynamic, competitive, and multi-agent interactions. In this work, we present a hierarchical multi-agent reinforcement learning (MARL) framework that enables fully autonomous and decentralized quadruped robot soccer. First, a set of highly dynamic low-level skills is trained for legged locomotion and ball manipulation, such as walking, dribbling, and kicking. On top of these, a high-level strategic planning policy is trained with Multi-Agent Proximal Policy Optimization (MAPPO) via Fictitious Self-Play (FSP). This learning framework allows agents to adapt to diverse opponent strategies and gives rise to sophisticated team behaviors, including coordinated passing, interception, and dynamic role allocation. With an extensive ablation study, the proposed learning method shows significant advantages in the cooperative and competitive multi-agent soccer game. We deploy the learned policies to real quadruped robots relying solely on onboard proprioception and decentralized localization, with the resulting system supporting autonomous robot-robot and robot-human soccer matches on indoor and outdoor soccer courts. 

\end{abstract}


\section{Introduction}
\label{sec:introduction}

Recent advances in deep reinforcement learning (DRL) have significantly improved single-agent capabilities of legged robots, enabling them to perform complex behaviors such as agile locomotion~\citep{he2024agile, zhu2024robust, zhuang2023robot, cheng2024extreme}, dynamic manipulation~\citep{he2024learning, cheng2025rambo}, and long-horizon navigation~\citep{lee2024learning, chen2018deep}. 
Despite these successes, real-world robotic systems often require collaboration among multiple agents, which is particularly challenging for legged robots due to their nonlinear dynamics, high-dimensional control spaces, and the need for strategic reasoning over extended time horizons.

Robot soccer serves as a high-profile benchmark for such settings, combining real-time control, decentralized decision making, and long-horizon strategy. Yet, despite growing interest, most prior work either adopts rule-based pipelines~\citep{labiosa2024reinforcement}, focuses on simplified 1v1 games~\citep{haarnoja2024learning, tirumala2024learning}, or remains confined to simulation~\citep{liu2022motor, li2024marladona}. Deploying a fully learning-based, competitive and cooperative soccer system on real legged robots remains a challenge.
Two main difficulties persist in solving this challenge. First, legged robots have to perform precise motor skills like dribbling and kicking, while maintaining balance and reacting to the highly dynamic motion of the ball. These skills require high-frequency joint-level control and are particularly sensitive to contact dynamics. Even small errors can lead to instability, collisions, or loss of ball control. Second, coordinated team play requires long-horizon strategic reasoning under decentralized execution, which is crucial in real-world systems to ensure robustness. Without a centralized coordinator, each robot must infer the intentions of teammates and opponents in a dynamically changing environment, making coordination difficult.

In this work, we propose a hierarchical Multi-Agent Reinforcement Learning (MARL) framework that addresses these challenges and enables real-world cooperative and competitive soccer between autonomous quadruped robot teams. 
To simultaneously achieve stable locomotion and ball manipulation, the framework begins by training a set of low-level skills, including walking, dribbling, and kicking.
These skills require precise, high-frequency control and are used consistently across different game configurations.
Building on these reusable skills, we train a decentralized high-level policy that learns to compose the low-level skills based on egocentric observation, as illustrated in Fig.~\ref{fig:teaser}. 
These policies are trained through Fictitious Self-Play (FSP)~\citep{heinrich2015fictitious}, where agents are iteratively trained against a population of past opponent policies, using Multi-Agent Proximal Policy Optimization (MAPPO)~\citep{yu2022ppo}, enabling the emergence of long-horizon strategic behaviors such as passing, interception, and dynamic role assignment.

We demonstrate and analyze 1v1, 2v1 and 2v2 soccer matches in simulation, and deploy 1v1 and 2v1 configurations in the real world.
Crucially, in the real-world deployment, each robot relies solely on its onboard sensors (including a LiDAR) for perception, enabling reliable ball detection in all directions even under poor lighting conditions. This design, which forgoes any external motion capture or centralized planner, yields fully decentralized, coordinated multi-agent soccer on real legged robots. Our results show that complex cooperative and competitive behaviors can emerge purely through learning, and the policy can be zero-shot transferred to physical robots.

Our main contributions are as follows. (1) We propose a novel hierarchical framework that composes multiple legged locomotion skills and high-level strategic planning, enabling fully learned cooperative and competitive behaviors for robot soccer. The use of FSP facilitates strategic policy evolution in adversarial settings. (2) We provide a systematic analysis of emergent multi-agent behaviors, highlighting key design choices that drive coordination and competition. (3) We demonstrate, for the first time, a fully decentralized multi-quadruped robotic system capable of playing soccer in the real world, supporting both robot-robot and robot-human games without external hardware infrastructure.


\section{Related Work}
\label{sec:related_work}
Legged robotic soccer originated in RoboCup~\citep{kitano1998robocup} challenges in the 1990s, with early teams generally relying on manually crafted gaits, vision routines, and rule-based tactics~\citep{browning2005stp, behnke2007hierarchical, yi2016hierarchical}. ~\citet{kohl2024policy} were among the first to show that learned locomotion policies could outperform manual tuning in robot soccer, a result that has been validated by many subsequent studies~\citep{haarnoja2024learning, schwab2018}. In the years since, learning-based control has steadily advanced the capabilities of legged robots in soccer. Many works have focused on single-robot skills, such as dribbling~\citep{ji2023dribblebot, hu2024dexdribbler, zhu2025dynamic}, kicking~\citep{ji2022hierarchical}, and goalkeeping~\citep{huang2023creating} to create a robust set of low-level skills essential for team gameplay.

By contrast, recent works in multi-agent robotic soccer gameplay remain limited. Since soccer is inherently a multi-agent game, MARL provides a strong algorithmic foundation to create cooperative and competitive robot teams. Among them, Centralized-Training-Decentralized-Execution (CTDE) frameworks such as MAPPO~\citep{yu2022ppo} have shown super-human coordination in soccer benchmarks, enabling emergent passing, zone defense, and role assignment in simulation~\citep{liu2022motor,li2024marladona}. For example, ~\citet{kim2021} proposed a two-stage centralized training scheme for heterogeneous 5v5 teams; ~\citet{abreu2025designing} scaled to 11-agent teams with some basic collaborations; ~\citet{liu2022motor} trained a 2v2 legged soccer game with emergent role allocation;  and~\citet{li2024marladona} introduced \textit{MARLadona}, a decentralized MARL framework that learns complex team behaviors in simulation. 

Yet, real-world deployments remain scarce, mainly limited to 1v1 encounters under controlled sensing. \citet{haarnoja2024learning} demonstrated state-based 1v1 bipedal matches using self-play RL, and~\citet{tirumala2024learning} showed similar capabilities with fully vision-based policies. To address the behavioral complexity of an increasing number of agents, works~\citep{labiosa2024reinforcement, browning2005stp} have adopted hierarchical frameworks that decompose tasks into reusable sub-policies, improving sample efficiency, and enabling flexible behavior switching. Among them,~\citet{labiosa2024reinforcement} demonstrated a 5v5 sim-to-real robot soccer system. However, these hierarchical approaches mainly rely on condition-specific manually crafted high-level strategies, requiring expert heuristics and intensive tuning. In comparison, learning-based approaches, which are capable of self-improving through autonomous gameplay, have not been demonstrated in real-world deployment. 

To date, real-world deployment of cooperative and competitive legged robot soccer leveraging autonomous learning-based control has not been demonstrated, a gap we aim to address in this work. 
	

\section{Method}
\label{sec:method}

In this section, we introduce a hierarchical framework that progressively learns low-level skills and high-level strategies in cooperative and competitive soccer scenarios. Our goal is to develop a modular learning framework for multi-agent quadruped soccer to produce coordinated and strategic behaviors both in simulation and on real-world robots. We posit that separating motor control from decision-making enables more effective learning and transfer, and that co-evolution through adversarial training drives the emergence of team strategies. We detail the design of the low-level and high-level policies in Sec.~\ref{subsec:low-level-skills} and Sec.~\ref{subsec:high-level-strategy}, and outline the training procedure for high-level strategies via FSP in Sec.~\ref{subsec: self-play}.

\begin{figure}[t]
	\vspace{0.0cm}
	\centering
	\begin{center}
		\includegraphics[width=\columnwidth]{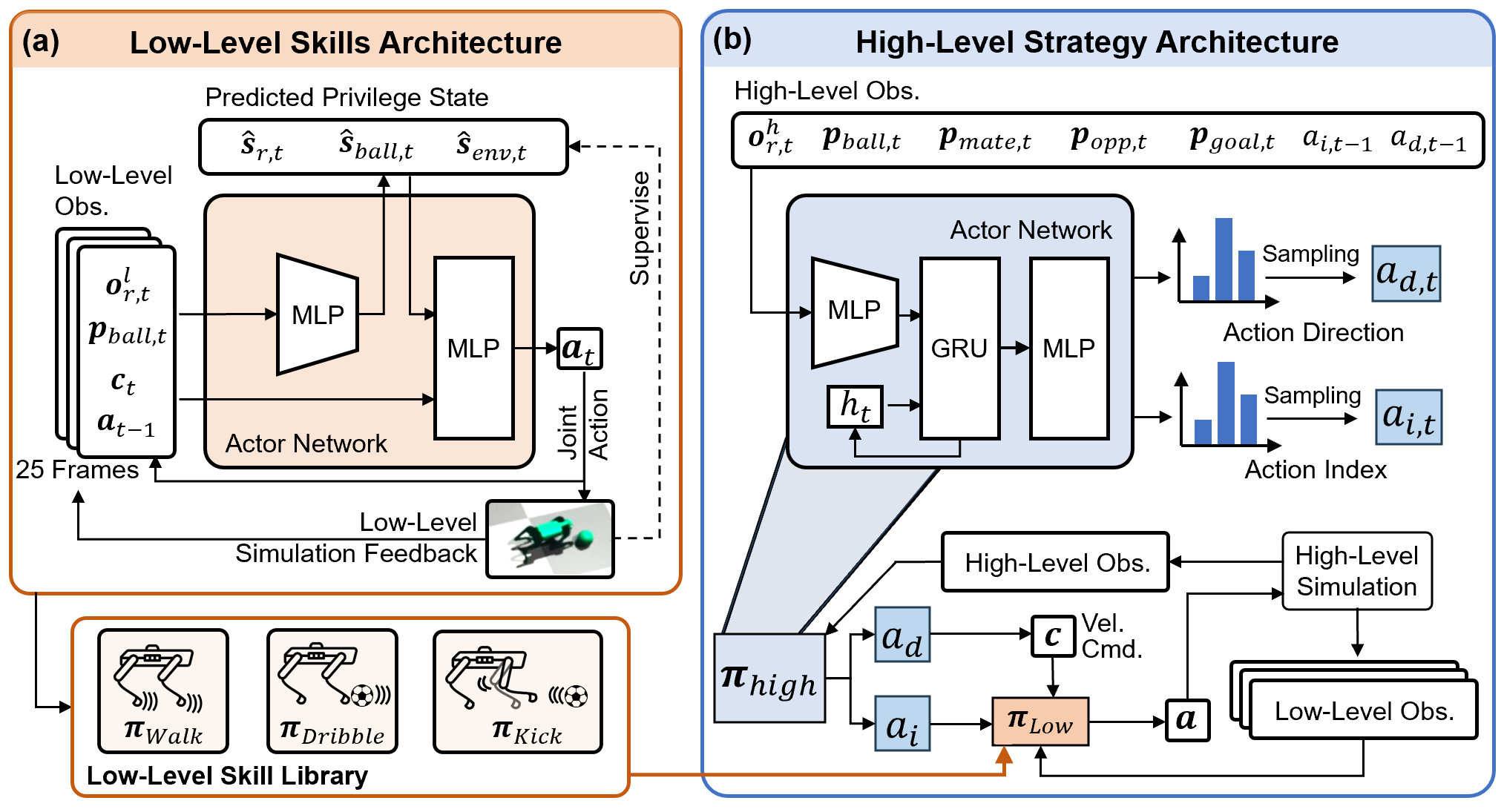}
	\end{center}
	\vspace{-0.4cm}
    \caption{Hierarchical Architecture. (a) \textbf{Low-level skills architecture}. The low-level observation is formed by concatenating the low-level robot observations $\boldsymbol{o}^l_{r,t}$, ball position $\boldsymbol{p}_{\text{ball},t}$, velocity command $\boldsymbol{c}_t$, and previous joint action $\boldsymbol{a}_{t-1}$. An auxiliary estimator network predicts privileged information unavailable to the policy, including the robot state $\boldsymbol{\hat{s}}_{r,t}$ (e.g., motor stiffness), ball state $\boldsymbol{\hat{s}}_{\text{ball},t}$ (e.g., ball velocity), and environment state $\boldsymbol{\hat{s}}_{\text{env},t}$ (e.g., friction coefficient), supervised using ground-truth data from the simulator. (b) \textbf{High-level strategy architecture}. A Gated Recurrent Unit (GRU) to maintain a hidden state $\boldsymbol{h}_{t}$ for long-term memory. The high-level actor will output discrete action index $a_i$ that selects the low-level skill type, and $a_d$ that determines the velocity command input to the chosen skill. These high-level actions are then concatenated with other observations and passed into the low-level policy for final execution. Check Appendix~\ref{app:training-details} for the observation space.}
	\label{fig:2layer}
	\vspace{-0.4cm}
\end{figure}

\subsection{Overview}

We study quadruped robotic soccer in a mixed cooperative-competitive setting, where robots are divided into \textbf{\textcolor{blue}{Attackers}} tasked with scoring goals and \textbf{\textcolor{red}{Defenders}} aiming to defend and counterattack.
This adversarial interaction between teams requires both intra-team coordination and inter-team competition. 
The resulting problem exhibits long horizons, sparse rewards, and high-dimensional continuous control, making direct joint-space reinforcement learning particularly challenging. 

To address these challenges, we adopt a hierarchical framework shown in Fig.~\ref{fig:2layer}, where the soccer policy is decoupled into
\begin{enumerate}[topsep=0pt, itemsep=0pt, parsep=0pt]
    \item[(i)] a library of reusable low-level motor skills learned separately, and
    \item[(ii)] a high-level scheduler that composes these skills online.
\end{enumerate}

This hierarchical structure (1) reduces the exploration burden, (2) yields interpretable behaviors, and (3) enables low-level skills to be transferred to different team configurations without re-training.

We primarily focus on the 2v1 setting for detailed study and analysis, while also validating our approach in the 1v1 and 2v2 configuration.
At the beginning of each game, the ball is placed near the primary attacker robot on its half of the court. 
A game terminates upon a goal, ball out of bounds, or timeout. A restricted zone is defined along all field boundaries, extending 0.5 meters inward from the edges. Robots are prohibited from entering these zones to prevent them from running out of bounds. All policies are trained in the GPU-accelerated simulator IsaacGym~\citep{makoviychuk2021isaac}, and selected policies are deployed zero-shot on real-world Go1 quadruped platforms~\citep{unitree2025website}.

\subsection{Low-Level Skill Control Policies}
\label{subsec:low-level-skills}
To enable the agile and robust behaviors necessary for soccer play, we first train a set of low-level skills to achieve dynamic locomotion and ball interaction.
Inspired by common maneuvers in human soccer, we developed three low-level skills, \textit{Walk}, \textit{Dribble}, and \textit{Kick}, which provide stable, interpretable building blocks that simplify the subsequent high‐level coordination problem. Training these low-level skills separately reduces exploration complexity and accelerates convergence, yielding skills that generalize across different team configurations and form a reliable foundation for strategic policy learning. The skills share a unified two-dimensional velocity command $\boldsymbol{c}=(v_x,v_y)$ defined in the world frame. 
For \textit{Walk}, $\boldsymbol{c}$ specifies the desired base velocity, and the robot is trained to align its yaw angle with the velocity direction. For \textit{Dribble}, $\boldsymbol{c}$ represents the target velocity of the ball. For \textit{Kick}, $\boldsymbol{c}$ is a unit vector that determines the desired velocity direction of the ball after kicking.
Each skill is realized by a neural network policy $\boldsymbol{\pi}_{\theta}(\boldsymbol{a}\,|\,\boldsymbol{o},\boldsymbol{c})$ which outputs desired joint positions at \(50\,\mathrm{Hz}\).
These target joint positions are subsequently tracked by proportional–derivative (PD) controllers operating at \(200\,\mathrm{Hz}\) with gains \(K_p=35\), \(K_d=0.5\).

We employ a model-free RL training algorithm Proximal Policy Optimization (PPO)~\citep{schulman2017proximal} to train those skills. Reward shaping follows prior work for \textit{Walk}~\citep{margolis2022walktheseways} and \textit{Dribble}~\citep{hu2024dexdribbler}. For \textit{Kick}, we introduce a state-conditioned multi-stage reward design. Refer to Appendix~\ref{app:training-details} for more details.

\subsection{High‑Level Multi-Agent Strategy Policy}
\label{subsec:high-level-strategy}

\begin{figure}[H]
  \centering
  \begin{minipage}[!h]{0.52\linewidth}
    With the skill library in place, we next explore how composing pre-trained soccer skills can yield adaptive, coordinated behaviors in dynamic soccer scenarios. A high-level controller composes these skills to play soccer in a cooperative and competitive manner, selecting a skill type with its 2D command. 
    To improve training efficiency, we discretize the continuous command $\boldsymbol{c}=(v_x,v_y)$ into eight equally spaced unit vectors. The command magnitude \(\|\boldsymbol{c}\|\) is pre‑specified per skill (e.g.\ \(v_\text{walk}\) for \emph{Walk}). In addition to the three skills introduced in Section~\ref{subsec:low-level-skills}, we include a \emph{Stop} action that holds the robot in place, regardless of the command. Table~\ref{tab:highlevel-action-space} summarizes the high‑level action space.
  \end{minipage}
  \hfill
    \begin{minipage}[!h]{0.47\linewidth}
      \centering
      \vspace{-1.0em}
      \captionof{table}{High-Level Policy Action Space}
      \label{tab:highlevel-action-space}
      \vspace{0.8em}
      \renewcommand{\arraystretch}{1.45}
        \begin{tabular}{p{0.7cm} c p{3.2cm}}
          \toprule
          \textbf{Skill Type} & & \textbf{Direction Options} \\
          \midrule
          Walk    & \multirow{3}{*}{\Bigg\} } &
                   \multirow{3}{*}{\parbox{3.2cm}{
                   \includegraphics[width=3.4cm]{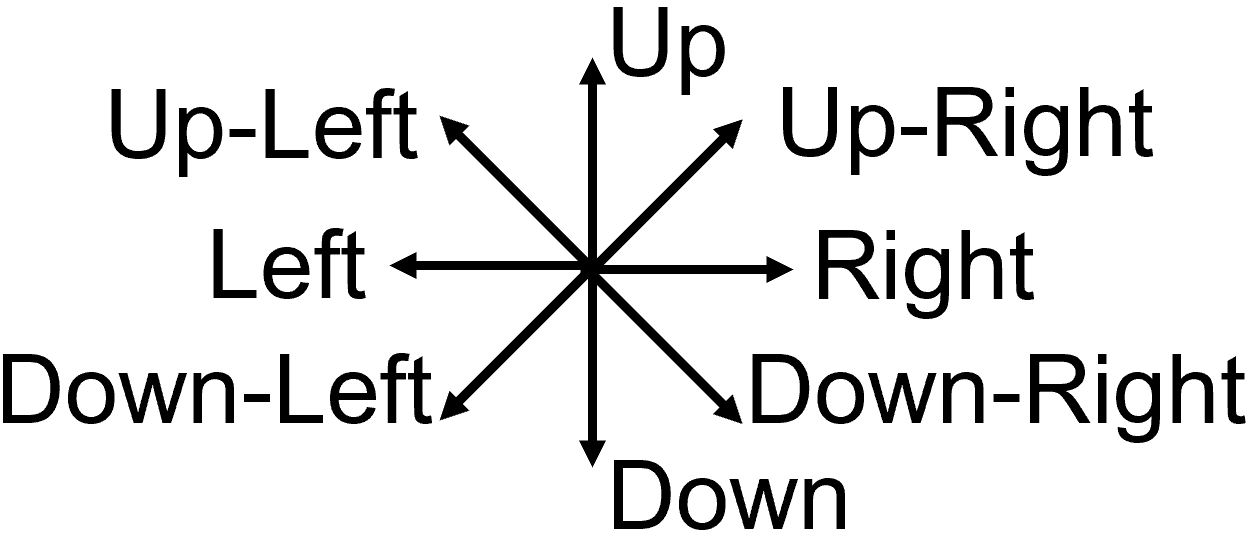}
                   }} \\
          Dribble & & \\
          Kick    & & \\
          \midrule
          Stop    & -- & None \\
          \bottomrule
        \end{tabular}
        
    \end{minipage}
    \vspace{-1.0em}
\end{figure}

The high-level policy receives only the current proprioceptive state and egocentric relative positions of teammates, opponents, the ball, and the goals on both sides. To capture long‑term dependencies, we implement the high‑level policy with a GRU~\citep{cho2014learning} backbone. Decisions are made every 10 low‑level steps (0.2 s, i.e.\ 5 Hz). This interval balances responsiveness with training efficiency. Two Softmax heads independently parameterize categorical distributions over primitives and directions. All agents on the same team share a common high-level policy network.

The reward structure combines sparse outcomes with auxiliary dense shaping. Agents receive a positive reward for scoring and a negative reward for conceding a goal. Dense rewards are used to encourage safe and purposeful behaviors, including minimizing collisions between robots, avoiding exiting the field bounds, and promoting forward ball progression. Importantly, no explicit reward is given for coordinated behaviors among robots; all cooperation arises implicitly through task-driven pressures, such as the need to score while avoiding interception by the opponent.

\subsection{Co-Evolution of Teams via Fictitious Self-Play}
\label{subsec: self-play}

\noindent
\begin{minipage}[t]{0.37\textwidth}
    Robot soccer is a typical mixed cooperative-competitive task, where the performance of the focal team is strongly coupled with that of the opponent team. 
    To enable efficient high-level strategy learning, it is crucial to have a well-matched opponent that continually challenges the focal team and promotes the emergence of sophisticated strategies. 
    Therefore, we introduce an FSP framework, in which attacker and defender teams are trained alternately to encourage co-evolution and strategic refinement.
\end{minipage}
\hfill
\begin{minipage}[t]{0.6\textwidth}
    \vspace{-10pt}
    \centering
    \includegraphics[width=0.99\linewidth,valign=t]{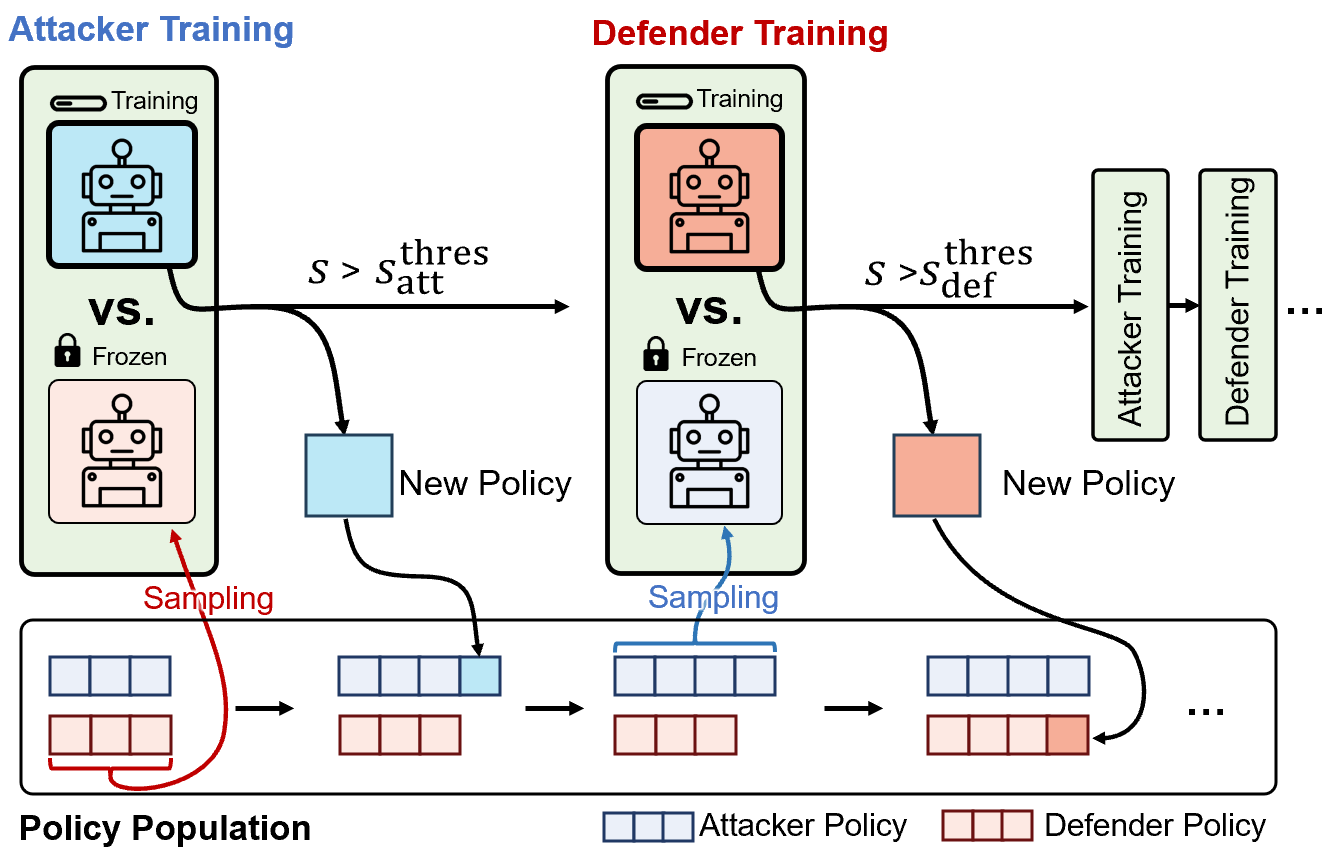} \\ 
    \captionof{figure}{The FSP training procedure, where each side is trained against a population of previously trained opponents.}
    \label{fig:self-play}
\end{minipage}

Specifically, attacker and defender policies are trained using MAPPO under an FSP regime: while one side updates, the opponent population pool is frozen. Training begins with the attackers, where the initial defender population consists of random and ball-chasing policies. The ball-chasing policy is a manually crafted high-level policy in which the robots continuously walk toward the ball. After each evaluation phase we compute the score
\begin{equation} s = r_{\text{win}} + 0.5 \times r_{\text{draw}}, \label{eq:score} \end{equation}
where $r_{\text{win}}$ and $r_\text{draw}$ denote the win and draw rates of the current training team, respectively. 
Whenever the focal team's score satisfies $s\geq s^{\text{thres}}$ (a pre-defined threshold, $s^{\text{thres}}_{\text{att}}$ for attacker and $s^{\text{thres}}_{\text{def}}$ for defender), the current policy snapshot is added to the population, and the training process switches to the opposing side. 
During focal team training, each environment independently and uniformly samples an opponent policy from the population. 
This curriculum promotes continuous adaptation while avoiding the instability of simultaneous updates.


\section{Experimental Results}
\label{sec:result}

In this section, we present comprehensive experimental results using the proposed method, including an ablation study, analyses of diverse emergent behaviors, and real-world deployment. Through our experiments in both simulation and the real world, we aim to answer the following questions: \textbf{Q1}: How does our hierarchical framework and the selection of three low-level skills influence the soccer training process? \textbf{Q2}: How does the FSP method induce diverse emergent behaviors? \textbf{Q3}: Can the learned value function provide insights into the cooperative and competitive behaviors of the high-level policy?

\subsection{Ablation Study}

\begin{figure}[h]
	\vspace{0.0cm}
	\centering
	\begin{center}
		\includegraphics[width=1.0\columnwidth]{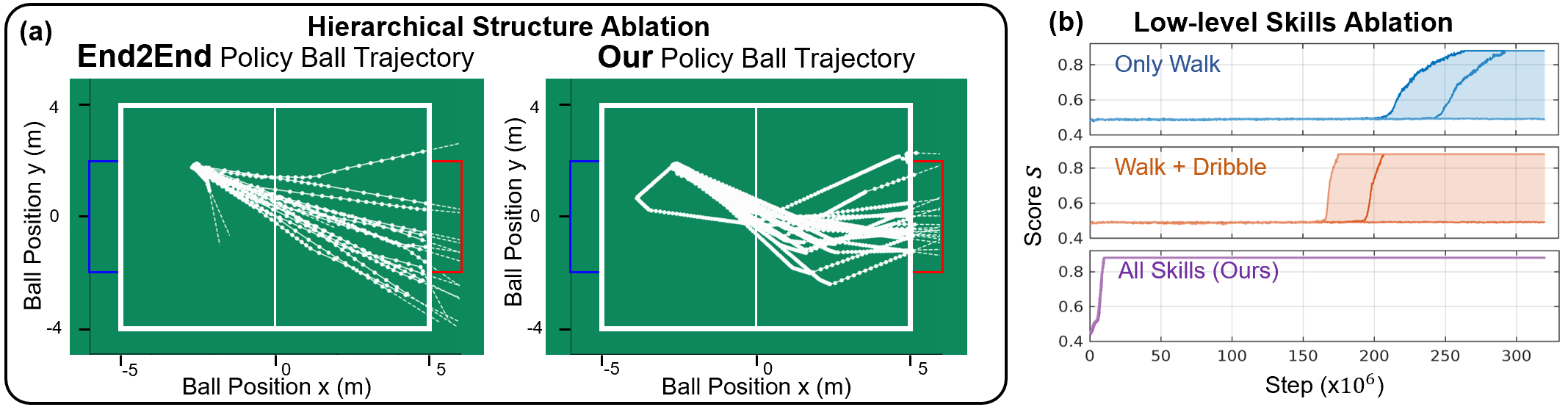}
	\end{center}
	\vspace{-0.2cm}
    \caption{Ablation Study. (a) Ball trajectories under \textbf{End2End} and \textbf{Ours} policies in simulation for 20 trails. (b) Training score $s$ curves comparing \textbf{Walk}, \textbf{Walk+Dribble}, and \textbf{Ours}, across three random seeds with the performance envelope colored in shaded area. We highlight that \textbf{Walk} and \textbf{Walk+Dribble} have substantially higher training variance, and even fail to explore any effective strategy after sufficient training under some random seeds.
    }
	\label{fig:ablation}
	\vspace{-0.2cm}
\end{figure}

We compare our hierarchical method (\textbf{Ours}) with several baselines, analyzing both behaviors and training efficiency.

First, we compare \textbf{Ours} with a flat end-to-end policy (\textbf{End2End}) that directly learns from proprioceptive observations to specify motor targets at 50 Hz. They are both trained against a stationary opponent for a sufficient duration. As shown in Fig.~\ref{fig:ablation}(a), the \textbf{End2End} policy exhibits unstable and uncoordinated ball control, often leading to erratic ball motions and frequent out-of-bounds events. In contrast, \textbf{Ours} achieves smooth and decisive ball trajectories toward the goal, demonstrating the effectiveness of skill abstraction in producing stable behaviors. Quantitatively, playing against a static opponent for 1000 episodes, \textbf{Ours} attains a win rate of 98.3\%, compared to 37.5\% for \textbf{End2End}, presenting better performance due to the hierarchical framework.

Second, we ablate the contribution of low-level skills by comparing three hierarchical variants: \textbf{Walk} (only walking skill), \textbf{Walk+Dribble} (walking and dribbling skills), and \textbf{Ours} (walking, dribbling, and kicking skills). All models are trained against a defender pool consisting of random and ball-chasing policies, and training stops once the attacker reaches a predefined score threshold ($s^{\text{thres}}_{\text{att}}=0.88$). As shown in Fig.~\ref{fig:ablation}(b), \textbf{Walk+Dribble} improves over \textbf{Walk} by enhancing ball control, but the lack of a kicking skill severely hampers exploration. Without the kicking skill, some runs fail to achieve any scoring success. \textbf{Ours} converges significantly faster, highlighting the importance of skill diversity for sample-efficient learning.

\subsection{Policy Evolution with Fictitious Self-Play}

\begin{figure}[t]
	\vspace{0.0cm}
	\centering
	\begin{center}
		\includegraphics[width=1.0\columnwidth]{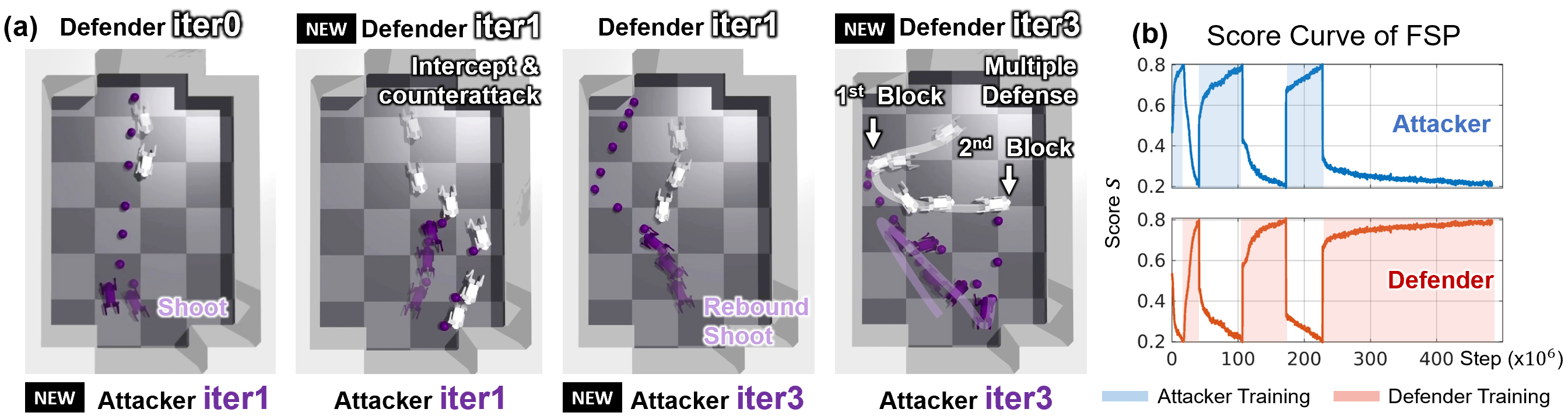}
	\end{center}
	\vspace{-0.2cm}
    \caption{Policy evolution with FSP in 1v1 setting.
    (a) Co-evolutionary dynamics between attacker and defender agents. Note that iter$n$ represents the policy trained after $n$ iterations.
    (b) Score $s$ (defined in Eq.~\ref{eq:score}) curve of FSP showing alternating improvements ( $s^{\text{thres}}_{\text{att}}=s^{\text{thres}}_{\text{def}}=0.8$).}
	\label{fig:1v1}
	\vspace{-0.1cm}
\end{figure}

We analyze how the FSP training induces diverse emergent behaviors of the high-level strategy.

Through FSP training, the attacker can gradually evolves from naive shooting to sophisticated tactics such as rebounding and adaptive shooting as the defender improves. Meanwhile, the defender also learns to perform dynamic interceptions and multi-stage defense over iterations, reflecting co-evolution between competing sides. 
The evolution of policy in the 1v1 setting is shown in Fig.~\ref{fig:1v1}.

\begin{figure}
	\vspace{0.0cm}
	\centering
	\begin{center}
		\includegraphics[width=1.0\columnwidth]{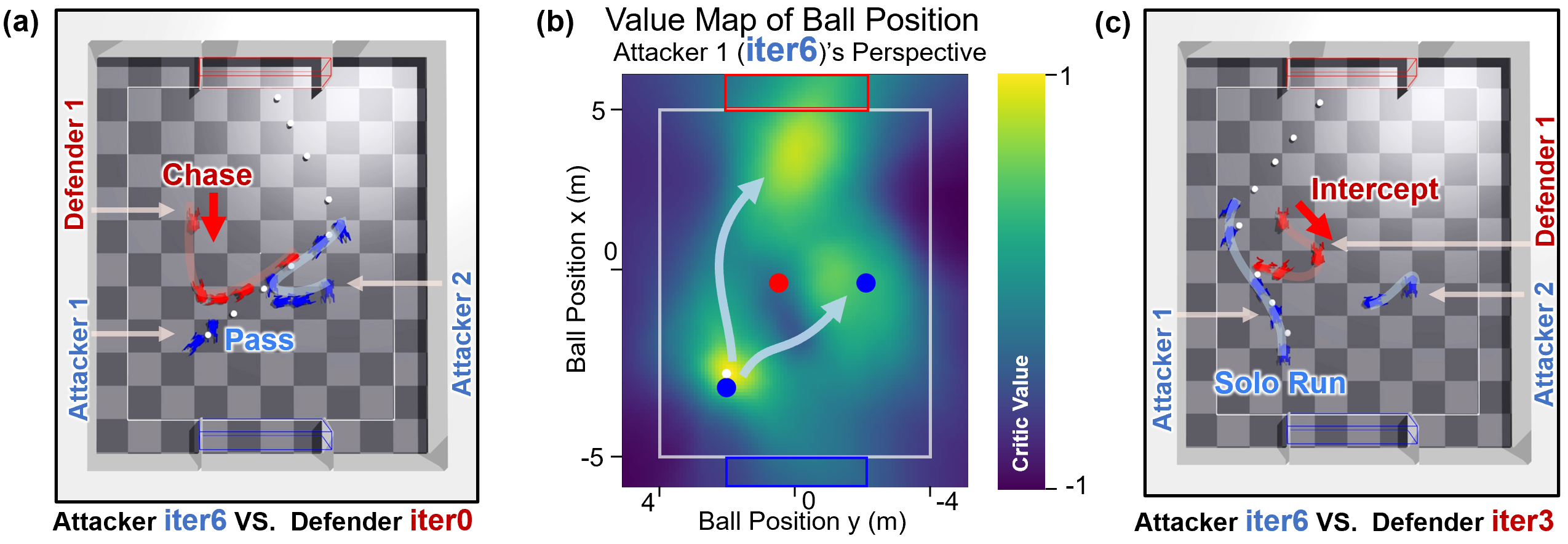}
	\end{center}
	\vspace{-0.4cm}
    \caption{
    Rollouts of policy trained with FSP in 2v1 setting.
    (b) For attackers, training against all previous defender snapshots yields a multi-modal distribution of diverse strategies without collapsing into single-modal local optima. 
    For instance, (a) and (c) illustrate two cases where the same attacker high-level policy exhibits different strategies when facing different defender policies.}
	\label{fig:2v1}
	\vspace{-0.4cm}
\end{figure}

Crucially, FSP enables the strategy policy to discover multi-modal behaviors and avoid local optima by training the attacker against a diverse population of past defender snapshots.  
For instance, in the 2v1 setting, the attacker learns to choose between passing and solo running depending on different defender's reactions and ball positions (Fig.~\ref{fig:2v1}).

These results highlight the effectiveness of FSP in driving behavioral evolution and strategy diversification. Additional emergent behaviors can be found in Appendix~\ref{app:more-strategies}. 
Our framework with FSP also scales effectively to 2v2 setting (Appendix~\ref{app:2v2}).

\subsection{Real-World Experiments Behavior Analysis with Value Function}

\begin{figure}[t]
	\vspace{0.0cm}
	\centering
	\begin{center}
		\includegraphics[width=1.0\columnwidth]{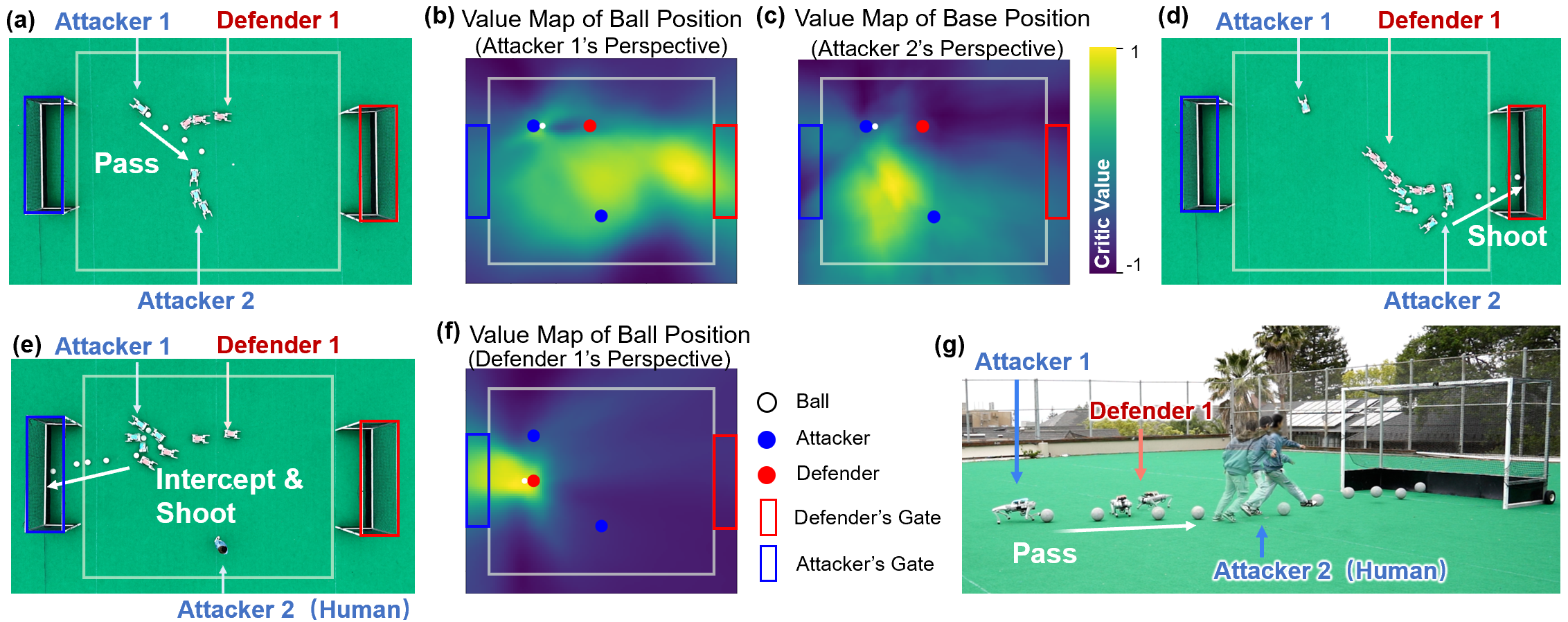}
	\end{center}
	\vspace{-0.3cm}
    \caption{Real-world behavior analysis. (a)(d) The top-down view of two attackers coordinating to pass and shoot against a defender. (b)(c) Critic value maps from the attackers’ perspectives show higher expected returns when Attacker 1 attempts a pass and Attacker 2 moves closer to receive. (e)(f) The defender intercepts the ball and counterattacks. The defender’s value map highlights favorable ball positions near the attackers’ goal and lower values near its own goal. (g) A robot attacker collaborates with a human teammate to score.}
	\label{fig:behavior-case-study}
	\vspace{-0.5cm}
\end{figure}

To build a decentralized multi-agent soccer system in the real world, we deploy our high- and low-level policies on three Unitree Go1 quadrupeds equipped with onboard sensors and computation, and local decision-making capabilities enabled by our decentralized policy architecture.
The system operates on a 10 m $\times$ 8 m field (Fig.\ref{fig:teaser}) without any external hardware infrastructure. 
Previous real-world robot soccer systems typically rely on external motion-capture systems~\citep{haarnoja2024learning}, which are impractical for outdoor deployment, or on vision-based tracking methods~\citep{tirumala2024learning}, which suffer from limited fields of view and vulnerability to motion blur.
In contrast, our robots achieve full autonomy using onboard LiDAR-based sensing (MID-360~\citep{livox2025website}) and onboard computation (NVIDIA Orin NX), enabling robust perception under fast motion, outdoor environments, and varying lighting. Localization is handled via real-time LiDAR-inertial odometry using FAST-LIO~\citep{xu2021fast} within a pre-mapped environment, allowing each robot to estimate its global pose independently. 
Only minimal pose information is broadcast among agents to enable mutual localization without external infrastructure.
This design enables fully autonomous, decentralized soccer play in both robot-robot and robot-human scenarios. Check Appendix~\ref{app:1v1} and~\ref{app:system} for indoor 1v1 real experiments and details of the real-world system. 

To further analyze the learned strategies in the real world, we visualize both the agents' behaviors and value maps (Fig.~\ref{fig:behavior-case-study}). In the first scenario shown in Fig.~\ref{fig:behavior-case-study}(a)–(d), the attackers are controlled by a trained high-level policy, competing against a defender with a handcrafted ball-chasing policy that always walks toward the ball. All low-level skill policies are shared by both sides. Against this reactive defender, the attackers’ value maps assign higher values to ball and base positions that favor passing the ball. As a result, the attackers coordinate through passing and successfully score.

In the second experiment (Fig.~\ref{fig:behavior-case-study}(e)–(g)), we replace the handcrafted ball-chasing high-level policy on Defender 1 with a trained one, and substitute Attacker 2 with a human teammate. This form of robot–human collaboration and competition is enabled by our fully decentralized control framework.
During the game, the defender successfully intercepts the ball and executes a counterattack, as reflected by its value map, which exhibits higher critic values near the attackers' goal and lower values near its own goal, aligning with its defensive objectives.
In another game, Attacker 1 successfully passes the ball to the human teammate, resulting in a goal.

In summary, these comprehensive experiments show that our hierarchical framework with all three low-level skills achieves the most stable and efficient learning, outperforming flat and ablated baselines. FSP promotes strategy diversification and mitigates local optima. The learned policy transfers zero-shot to the real world, with value maps revealing interpretable cooperative and competitive behaviors—together highlighting the strengths of our design choices.


\section{Conclusion}
\label{sec:conclusion}

In this work, we propose a hierarchical MARL framework to achieve decentralized quadruped soccer, combining learned low-level motor skills with a high-level multi-agent strategy. Our method outperforms flat and restricted baselines in both stability and sample efficiency. Through an FSP regime, our agents develop versatile competitive and cooperative behaviors such as passing, blocking, and counter-attacking. Real-world experiments with multiple Unitree Go1 robots validate the effectiveness of our approach on indoor and outdoor soccer courts, taking a step toward autonomous robot soccer teams that can rival a human team.


\section{Limitations}
\label{sec:limitation}

While our hierarchical MARL framework enables robust multi-agent coordination for quadruped soccer, several limitations remain. First, this work primarily focuses on 1v1 and 2v1 settings; scaling to larger team sizes as in prior works presents challenges for both efficient training and real-world deployment because there is a sharp increase in sample complexity and communication overhead as team sizes increase. Real-world deployment of large teams also exacerbates safety and synchronization concerns; in particular, our LiDAR-based ball detection may suffer from frequent occlusions as the number of robots increases. One promising direction is to simulate occlusion during training and encourage agents to actively infer and search for the ball when it is not directly observable. Additionally, our current experiments are limited to quadruped platforms. Extending the framework to full-size humanoid robots would introduce significant additional challenges in maintaining stable locomotion and achieving precise ball interactions.

\clearpage
\acknowledgments{This work is supported by the Robotics and AI Institute. Z. Su receives financial support from the Institute for Interdisciplinary Information Sciences (IIIS) at Tsinghua University, and Y. Gao is supported by Zhejiang University. C. Yu receives support from National Natural Science Foundation of China (No.62406159). The authors would like to thank Xiaoyu Huang and Qiayuan Liao for their assistance with the experiments. K. Sreenath has a financial interest in the Robotics and AI Institute. He and the company may benefit from the commercialization of the results of this research. This work was done when Z. Su and Y. Gao visited UC Berkeley.}


\bibliography{example}  

\newpage
\appendix
\section{Low-level Skills Demonstration}
\label{app:low-level-skills}
\begin{figure}[h]
	\vspace{0.0cm}
	\centering
	\begin{center}
		\includegraphics[width=1.0\columnwidth]{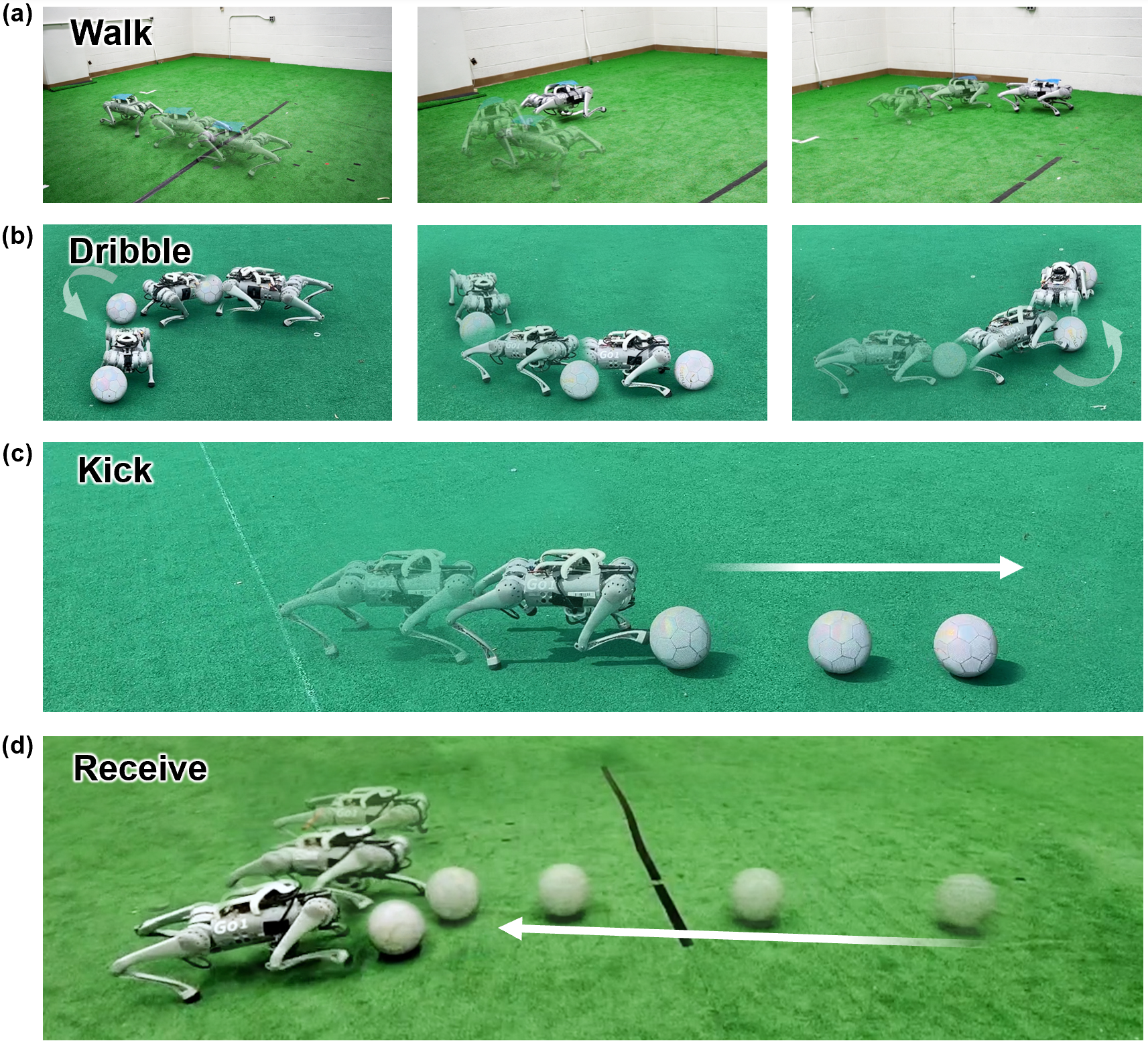}
	\end{center}
	\vspace{-0.3cm}
    \caption{Real-world demonstration of low-level skills: (a) \textit{Walk} (b) \textit{Dribble} (c) \textit{Kick} (d) \textit{Receive}.}
	\label{fig:real-low-level}
\end{figure}

We demonstrate the four low-level motor skills deployed on real quadruped robots.
Fig.~\ref{fig:real-low-level}(a) shows the \textit{Walk} primitive, which enables stable omnidirectional locomotion with the robot oriented in the direction of travel. Fig.~\ref{fig:real-low-level}(b) demonstrates the \textit{Dribble} skill, where the robot approaches the ball, maintains alignment, and guides it through repeated contact and reorientation. Fig.~\ref{fig:real-low-level}(c) presents the \textit{Kick} skill, where the robot moves toward the ball and delivers a powerful strike to propel it forward. These low-level skills form the foundation for composing higher-level strategic behaviors.

The \textit{Receive} skill illustrated in Fig.~\ref{fig:real-low-level}(d) allows the robot to stop and handle an incoming ball. While this skill is occasionally selected during early training stages, we observe that it is consistently abandoned after sufficient learning. This is because receiving the ball often incurs a significant delay: the robot must first turn to face the teammate to stop the ball, and then turn again to advance toward the opponent’s goal. In contrast, it is more efficient to directly exploit the ball's forward momentum by immediately dribbling and kicking without stopping. As a result, the high-level policy naturally and autonomously avoids selecting the \textit{Receive} skill. To improve training efficiency, we remove it from the final skills.

\section{Real-world Experiments for 1v1 Game}
\label{app:1v1}
\begin{figure}[h]
	\vspace{0.0cm}
	\centering
	\begin{center}
		\includegraphics[width=1.0\columnwidth]{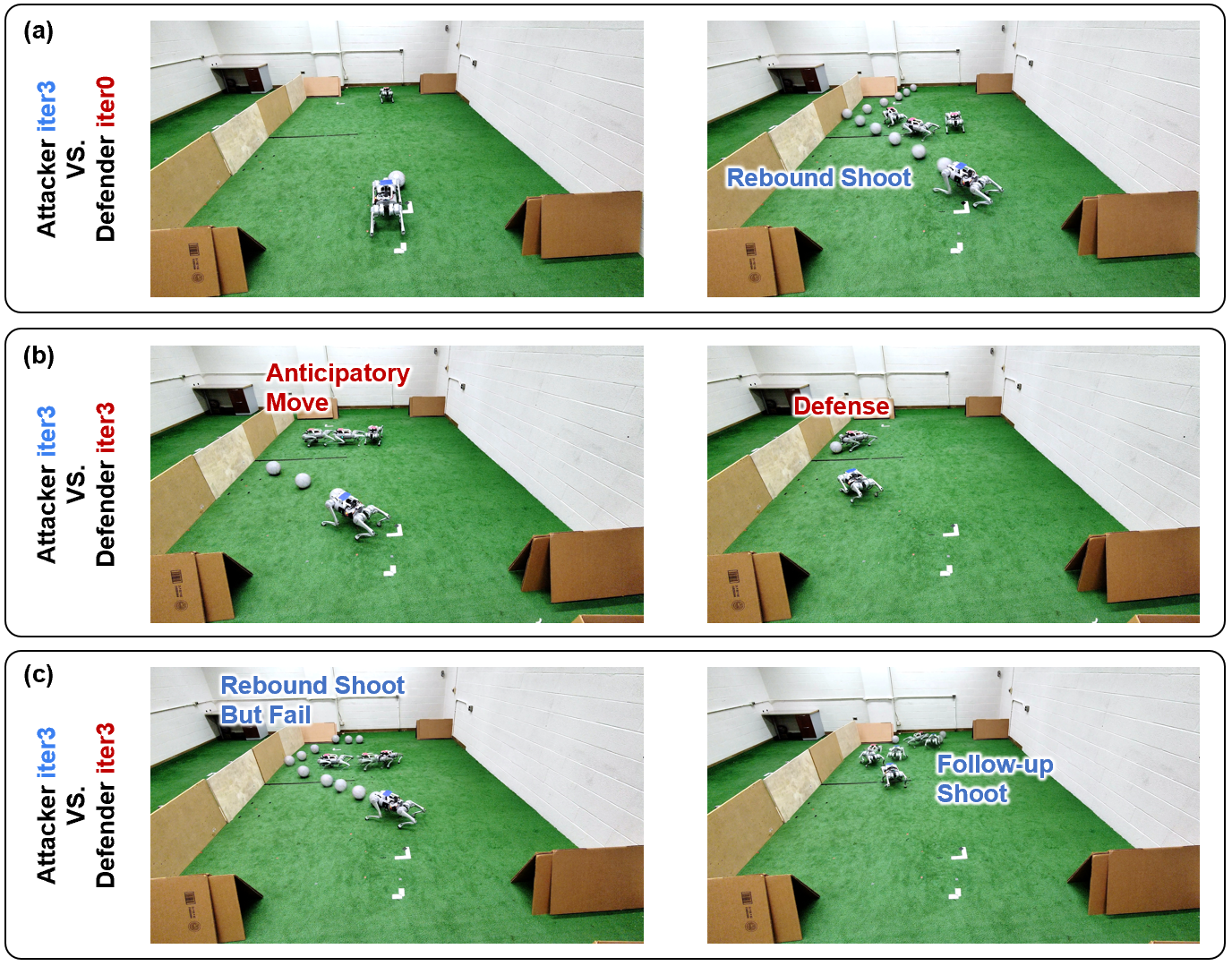}
	\end{center}
	\vspace{-0.3cm}
    \caption{Demonstration of 1v1 real world deployment.}
	\label{fig:1v1-real}
\end{figure}

We further deploy our trained 1v1 policies in the real world and showcase three representative cases in Fig.~\ref{fig:1v1-real}. After three iterations of training, the attacker learns to exploit the wall to rebound the ball into the goal (Fig.~\ref{fig:1v1-real}(a)). Simultaneously, the defender exhibits anticipatory behavior by moving proactively to intercept the rebounding ball, demonstrating co-evolved competitive dynamics (Fig.~\ref{fig:1v1-real}(b)). 
In another scenario, the attacker initially fails to score, but successfully scores by a follow-up shot, illustrating the robustness and persistence of the learned policy (Fig.~\ref{fig:1v1-real}(c)).

\section{More Emergent Strategies for 2v1 Game}
\label{app:more-strategies}
\begin{figure}[h]
	\vspace{0.0cm}
	\centering
	\begin{center}
		\includegraphics[width=0.8\columnwidth]{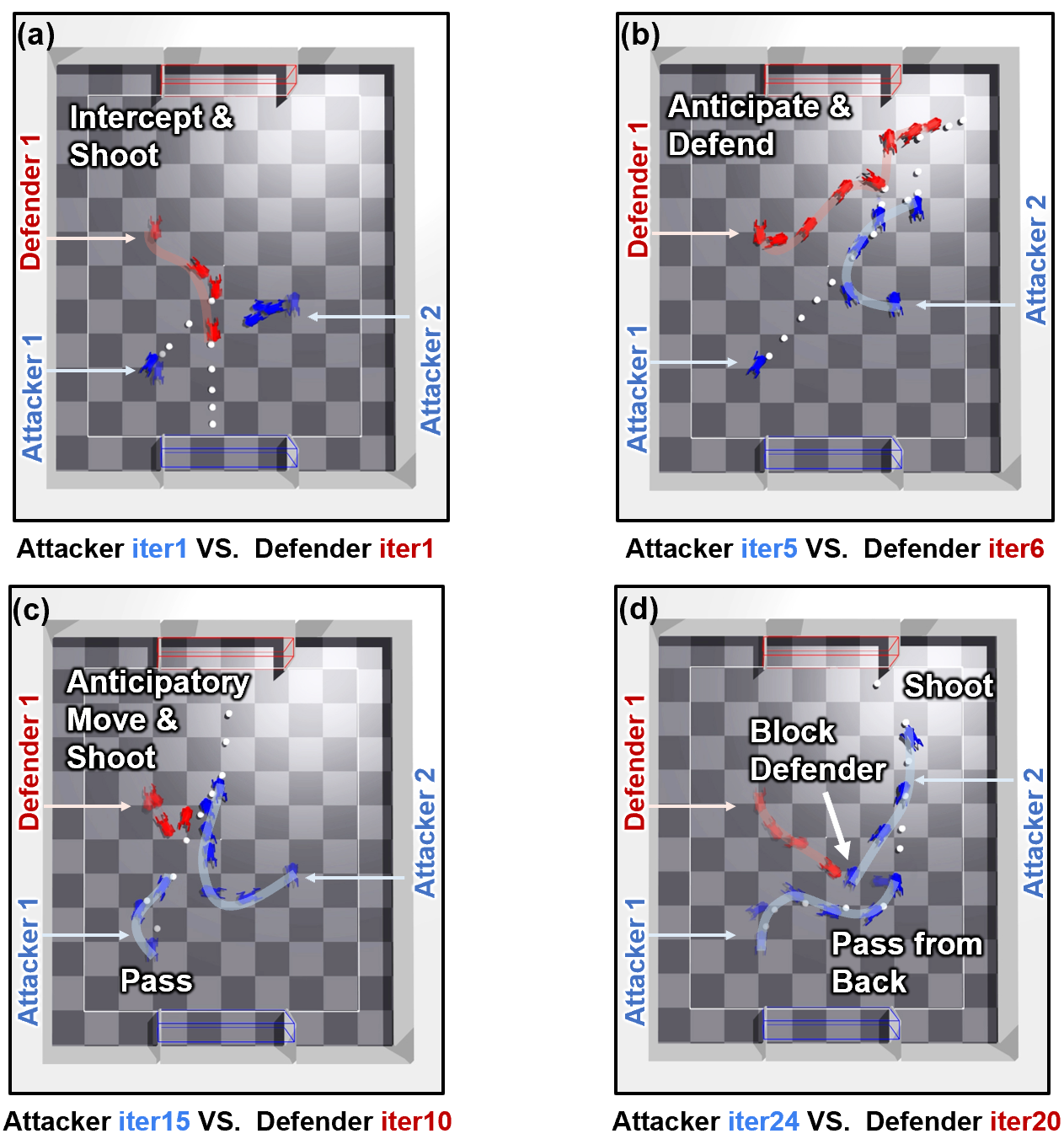}
	\end{center}
	\vspace{-0.3cm}
    \caption{Demonstration of more emergent high-level strategies in 2v1 game. (a)-(b): The defender performed effective defenses. (c)-(d): The attackers executed coordinated attacks. }
	\label{fig:more-2v1-behaviors}
\end{figure}

Here, we illustrate more emergent high-level strategies observed during training in the 2v1 setting.

In Fig.~\ref{fig:more-2v1-behaviors}(a), Defender 1 breaks up a pass between the attackers and counterattacks to score.

In Fig.~\ref{fig:more-2v1-behaviors}(b), Defender 1 anticipates a potential shot from Attacker 2 by proactively approaching its own goal, successfully intercepting two shot attempts.

In Fig.~\ref{fig:more-2v1-behaviors}(c), Attacker 1 waits patiently to attract the defender's attention, luring Defender 1 closer. Meanwhile, Attacker 2 moves in advance, receives the pass from Attacker 1, and finishes with a shot.

In Fig.~\ref{fig:more-2v1-behaviors}(d), a more sophisticated collaborative behavior emerges. Attacker 2 actively blocks Defender 1's path, enabling Attacker 1 to move behind them and delivers a forward pass from behind. Attacker 2 then receives the ball and completes the play with a shot on goal. Once this strategy is acquired, it effectively prevents the defender from gaining access to the ball.

\section{Training Details}
\label{app:training-details}
\subsection{Low-level Skills Training}
\paragraph{Partially Observable Markov Decision Process} 
We formulate the low-level control problem as a Partially Observable Markov Decision Process (POMDP) defined by a state space $\mathcal{S}$, an action space $\mathcal{A}$, a transition function $p(\boldsymbol{s}_{t+1} | \boldsymbol{s}_t, \boldsymbol{a}_t)$, and a reward function $r(\boldsymbol{s}_t, \boldsymbol{a}_t)$. At each timestep $t$, the agent receives an observation $\boldsymbol{o}_t$ that provides partial information about the true state $\boldsymbol{s}_t$. The policy $\pi(\boldsymbol{a}_t | \boldsymbol{x}_t)$ takes as input a history of the most recent $h$ observations, $\boldsymbol{x}_t = (\boldsymbol{o}_{t-h+1}, \boldsymbol{o}_{t-h+2}, \dots, \boldsymbol{o}_t)$, and outputs an action $\boldsymbol{a}_t$. The objective of the reinforcement learning algorithm is to learn an optimal policy $\pi^*$ that maximizes the expected cumulative discounted reward:

\[
\pi^* = \arg\max_{\pi} \mathbb{E}_{\pi} \left[ \sum_{t} \gamma^t r(\boldsymbol{s}_t, \boldsymbol{a}_t) \right],
\]

where $\gamma \in (0,1]$ is the discount factor.

\paragraph{Observations}
Our low-level policies take as input a sequence of 25 recent observation frames. Each frame consists of multiple proprioceptive and task-relevant features, as summarized in Table~\ref{tab:obs-features}.

\begin{table}[h]
  \centering
  \caption{Low-Level Policy Observation Terms per Frame}
  \label{tab:obs-features}
  \begin{tabular}{lllp{6cm}}
    \toprule
    \textbf{Observation Term} & \textbf{Dim.} & \textbf{Symbol} &\textbf{Description} \\
    \midrule
    \texttt{base\_lin\_vel} & 3 & \multirow{7}{*}{$\boldsymbol{o}^l_{r,t}$} & Base linear velocity in base frame \\
    \texttt{base\_ang\_vel} & 3 &  & Base angular velocity in base frame \\
    \texttt{forward\_vec} & 2 & &Heading vector of the robot in world frame \\
    \texttt{projected\_gravity} & 3 & & Gravity vector projected into base frame \\
    \texttt{dof\_pos} & 12 & & Joint positions \\
    \texttt{dof\_vel} & 12 & &Joint velocities \\
    \texttt{gait\_sin\_indict} & 4 & &Gait clock input signals \\
    \midrule
    \texttt{ball\_states\_p} & 3 & $\boldsymbol{p}_{\text{ball},t}$ &Relative ball position in base frame (only for dribbling and kicking) \\
    \texttt{command} & 2 & $\boldsymbol{c}_t$ &Velocity command in world frame \\
    \texttt{last\_actions} & 12 & $\boldsymbol{a}_{t-1}$ & Actions taken in previous step \\
    \bottomrule
  \end{tabular}
\end{table}

\paragraph{Kicking Reward Functions}
We carefully design a state-conditioned multi-staged reward function to train the kicking policy. We divide kicking behavior into two stages:
\begin{itemize}
  \item \textbf{Pursue\,\&\,Strike stage} (\(r^{\text{kick}}_t<r^{\text{thres}}\)). The robot is rewarded for walking behind the ball, facing the command direction, and executing a kick.
  \item \textbf{Hold stage} (\(r^{\text{kick}}_t\ge r^{\text{thres}}\)). Once $r^{\text{kick}}$ exceeds the threshold, indicating a satisfactory strike, the robot is rewarded for stabilizing its posture in place. Otherwise, it continues repositioning and attempts to kick again.
\end{itemize}
Here, $r^{\text{kick}}$ is a kick-quality reward function and \(r^{\text{thres}}\) is a fixed threshold. The overall reward is given by: $r=r^{\text{hold}}\times \mathbbm{1}(r^{\text{kick}}_t\ge r^{\text{thres}})+r^{\text{strike}}\times\mathbbm{1}(r^{\text{kick}}_t< r^{\text{thres}})$.

The main reward terms are listed in Table~\ref{tab:kick-rewards}. We omit standard regularization terms such as torque penalty and action smoothness for brevity.

\begin{table}[h]
  \centering
  \caption{Main Reward Terms for the Kicking Policy}
  \label{tab:kick-rewards}
  \begin{tabular}{lp{8cm}}
    \toprule
    \textbf{Reward Term} & \textbf{Description} \\
    \midrule
    \texttt{kicking\_ball\_vel} ($r_{\text{kick}}$) & Encourages ball velocity in the commanded direction \\
    \texttt{dribbling\_robot\_ball\_yaw} & Aligns the robot-ball and robot-yaw direction with the command vector \\
    \texttt{dribbling\_robot\_ball\_pos} & Encourages the robot to keep the ball close if $r_{\text{kick}} < r^{\text{thres}}$ \\
    \texttt{dribbling\_robot\_ball\_vel} & Encourages the robot to approach the ball if $r_{\text{kick}} < r^{\text{thres}}$ \\
    \texttt{tracking\_lin\_vel} & Encourages the robot to stay still if $r_{\text{kick}} \geq r^{\text{thres}}$ \\
    \bottomrule
  \end{tabular}
\end{table}

\subsection{High-level Strategy Training}
\paragraph{Decentralized Partially Observable Markov Game} 
We formulate the high-level decision problem as a Decentralized Partially Observable Markov Game (Dec-POMG) defined by a set of agents $\mathcal{I}$, an enviornment state space $\mathcal{S}$, a joint observation space of all agents $\mathcal{O}=\{\boldsymbol{o}^i\}_{i=1}^N$, a joint action space of all agents $\mathcal{A}=\{\boldsymbol{a}^i\}_{i=1}^N$, a transition function $p(\boldsymbol{s}_{t+1} | \boldsymbol{s}_t, \boldsymbol{a}^1_t, \dots, \boldsymbol{a}^N_t)$, and reward functions ${r_i(\boldsymbol{s}_t, \boldsymbol{a}^1_t, \dots, \boldsymbol{a}^N_t)}_{i=1}^N$ for $N$ agents. At each timestep $t$, agent $i$ receives an observation $\boldsymbol{o}^i_t$ that provides partial information about the true state $\boldsymbol{s}_t$, and selects an action $\boldsymbol{a}^i_t \sim \pi_i(\boldsymbol{a}^i_t|\boldsymbol{o}^i_t)$ based on its observation. The objective of each agent is to learn a policy $\pi_i^*$ that maximizes its own expected cumulative discounted reward:
\[
\pi_i^* = \arg\max_{\pi_i} \mathbb{E}_{\pi_1, \dots, \pi_N} \left[ \sum_{t} \gamma^t r_i(\boldsymbol{s}_t, \boldsymbol{a}^1_t, \dots, \boldsymbol{a}^N_t) \right],
\]
where $\gamma \in [0, 1)$ is the discount factor. 
For this paper, all agents on the same team share a common policy.

\paragraph{Observations}
Since the high-level policy uses a GRU to aggregate temporal context, it only requires a single-frame observation as input. The observation space only includes local features, as detailed in Table~\ref{tab:highlevel-obs}.

\begin{table}[h]
  \centering
  \caption{High-Level Policy Observations}
  \label{tab:highlevel-obs}
  \begin{tabular}{lllp{6cm}}
    \toprule
    \textbf{Observation Term} & \textbf{Dim.} & \textbf{Symbol} & \textbf{Description} \\
    \midrule
    \texttt{forward\_vec} & 2 & $\boldsymbol{o}^h_{r,t}$ & Heading vector of the robot in world frame \\
    \texttt{ball\_pos\_xy} & 2 & $\boldsymbol{p}_{\text{ball},t}$ & Relative position of the ball in base frame \\
    \texttt{teammate\_pos\_xy} & $2 \times (n_{\text{team}} - 1)$ &$\boldsymbol{p}_{\text{mate},t}$ &Relative positions of teammates in base frame \\
    \texttt{oppo\_pos\_xy} & $2 \times n_{\text{opp}}$ &$\boldsymbol{p}_{\text{opp},t}$ &Relative positions of opponents in base frame \\
    \midrule
    \texttt{oppo\_goal\_pos\_xy} & 2 &\multirow{2}{*}{$\boldsymbol{p}_{\text{goal},t}$} &Relative position of the opponent goal in base frame \\
    \texttt{self\_goal\_pos\_xy} & 2 & &Relative position of the agent's own goal in base frame \\
    \midrule
    \texttt{last\_action} & 2 &\makecell{$a_{i,t-1}$ \\ $a_{d,t-1}$} &High-level action at the previous decision step \\
    \bottomrule
  \end{tabular}
    \begin{tablenotes}
      \footnotesize
      \item $n_{\text{team}}$ refers to the number of robots on the agent’s own team, while $n_{\text{opp}}$ denotes the number of opponents.
    \end{tablenotes}
\end{table}

\paragraph{Reward Functions}
The high-level strategy is primarily guided by sparse event-based rewards, such as scoring and conceding, which capture the long-term objectives of the game. To facilitate more efficient learning, we additionally incorporate dense auxiliary rewards that encourage intermediate behaviors like ball advancement.

\begin{table}[h]
  \centering
  \caption{High-Level Reward Terms}
  \label{tab:highlevel-reward}
  \begin{tabular}{llp{8cm}}
    \toprule
    \textbf{Reward Term} & \textbf{Weight} & \textbf{Description} \\
    \midrule
    \texttt{scoring} & $1000.0$ & Reward for successfully scoring a goal \\
    \texttt{conceding} & $-1000.0$ & Penalty for conceding a goal \\
    \texttt{out\_of\_border} & $-500.0$ & Penalty when the ball goes out of bounds \\
    \texttt{ball\_forward\_pos} & $1.0$ & Reward for advancing the ball toward the opponent goal \\
    \texttt{ball\_forward\_vel} & $1.0$ & Reward for ball velocity in the forward direction \\
    \texttt{base2ball} & $0.3$ & Reward for approaching the ball when the agent is the closest teammate \\
    \texttt{interference} & $-3.0$ & Penalty for being too close to other robots \\
    \texttt{penalty\_area} & $-0.3$ & Penalty for being too close to the boundary (all positive rewards are disabled if robot is in the penalty area) \\
    \texttt{fall\_over} & $-5.0$ & Penalty for falling over \\
    \texttt{opponent\_near\_ball} & $-5.0$ & Penalty if an opponent is close to the ball \\
    \bottomrule
  \end{tabular}
\end{table}

\paragraph{Transition Rule between High-Level Actions and Low-Level Commands}
Before converting a high-level action into a low-level command, we apply the following rule-based transitions to ensure appropriate skill activation:

\begin{enumerate}
    \item \textbf{Dribbling} is only activated when the robot is sufficiently close to the ball; otherwise, it is mapped to a walking command directed toward the ball.
    \item \textbf{Kicking} is only activated when the robot is sufficiently close to the ball; otherwise, it is mapped to a stationary stepping.
    \item Once \textbf{kicking} is initiated at a high-level decision step, the robot continues executing the kick until the ball is determined to have moved far away, indicating a successful kick.
    \item \textbf{Walking} toward other robots is prevented by predicting future positions using the current location and velocity command, and checking for potential collisions.
\end{enumerate}

These transition rules are consistently applied during both training and real-world deployment.

\section{Out of Domain Test}
\label{app:ood-test}
\begin{figure}[h]
    \centering
    \includegraphics[width=0.9\linewidth]{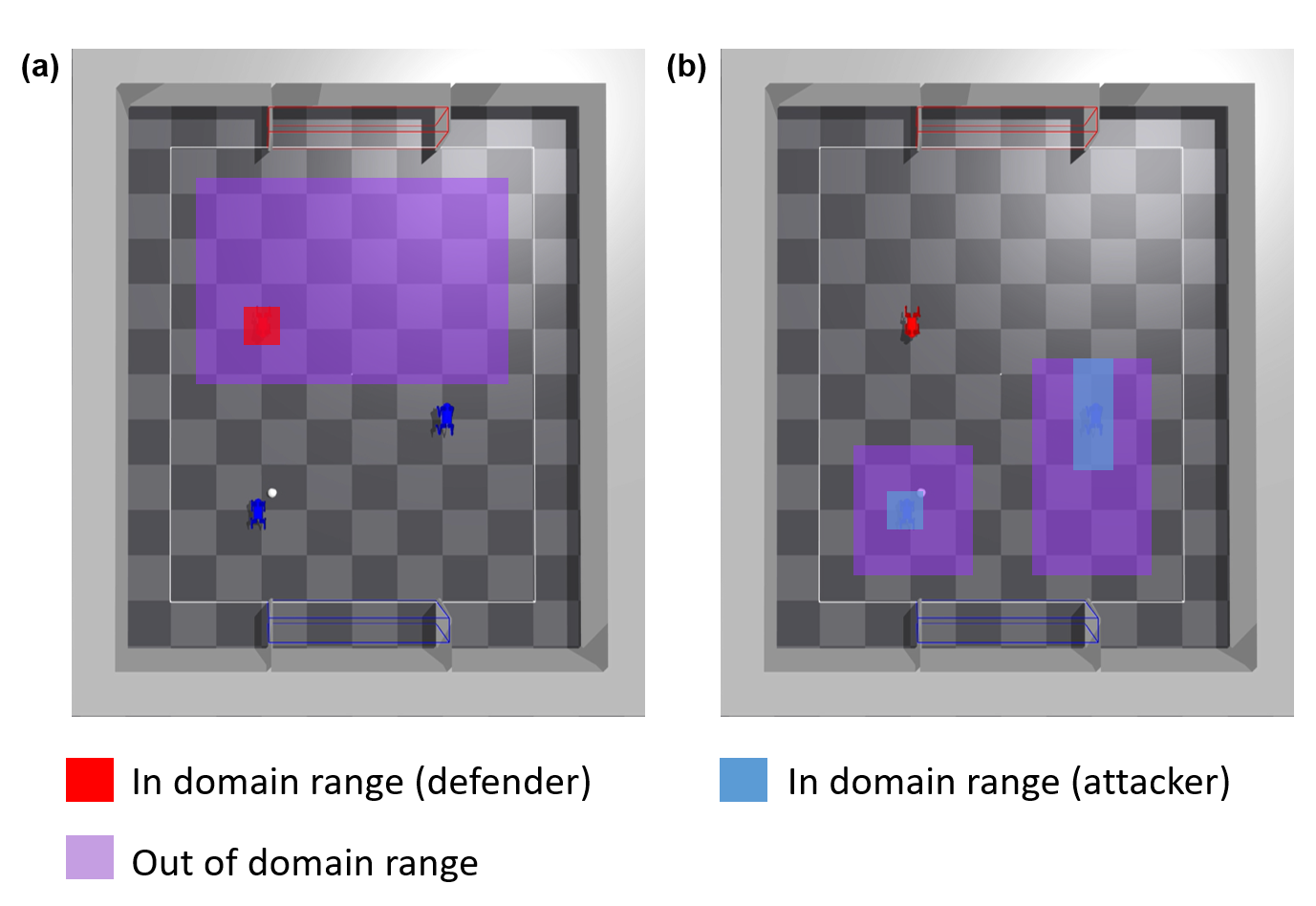}
    \caption{Visualization of the out-of-domain test setup. The highlighted rectangular regions illustrate the randomization ranges of the robots' initial positions. The red / blue patches corresponds to the in-domain regions used during training, while the purple patches represent the out-of-domain regions used for testing. The patches are extended slightly to account for the robots’ body length.}
    \label{fig:ood-test}
\end{figure}

To assess the robustness of our trained policy, we conduct additional out-of-domain evaluations by varying the robots’ initialization positions. In the 2v1 configuration, we evaluate the final attacker policy against a fixed ball-chasing defender under three settings, each consisting of 1000 episodes:
\begin{enumerate}
    \item In-domain: both the defender and attackers are initialized within the same small region used during training;
    \item Out-of-domain (defender position): the defender is initialized within a larger region on its side, while the attackers' initialization remains unchanged;
    \item Out-of-domain (attacker position): the attackers are initialized within a larger region on their side, while the defender's initialization remains unchanged (Fig.~\ref{fig:ood-test}).
\end{enumerate}

The win rate in the in-domain setting is 89.2\%. In the out-of-domain setting with a randomized defender position, the win rate remains high at 81.9\%. When the attackers are initialized out-of-domain, the win rate is 68.9\%. These results demonstrate the robustness and generalization capability of our trained attacker policy.

\section{2v2 Game in Simulation}
\label{app:2v2}
\begin{figure}[h]
    \centering
    \includegraphics[width=0.9\linewidth]{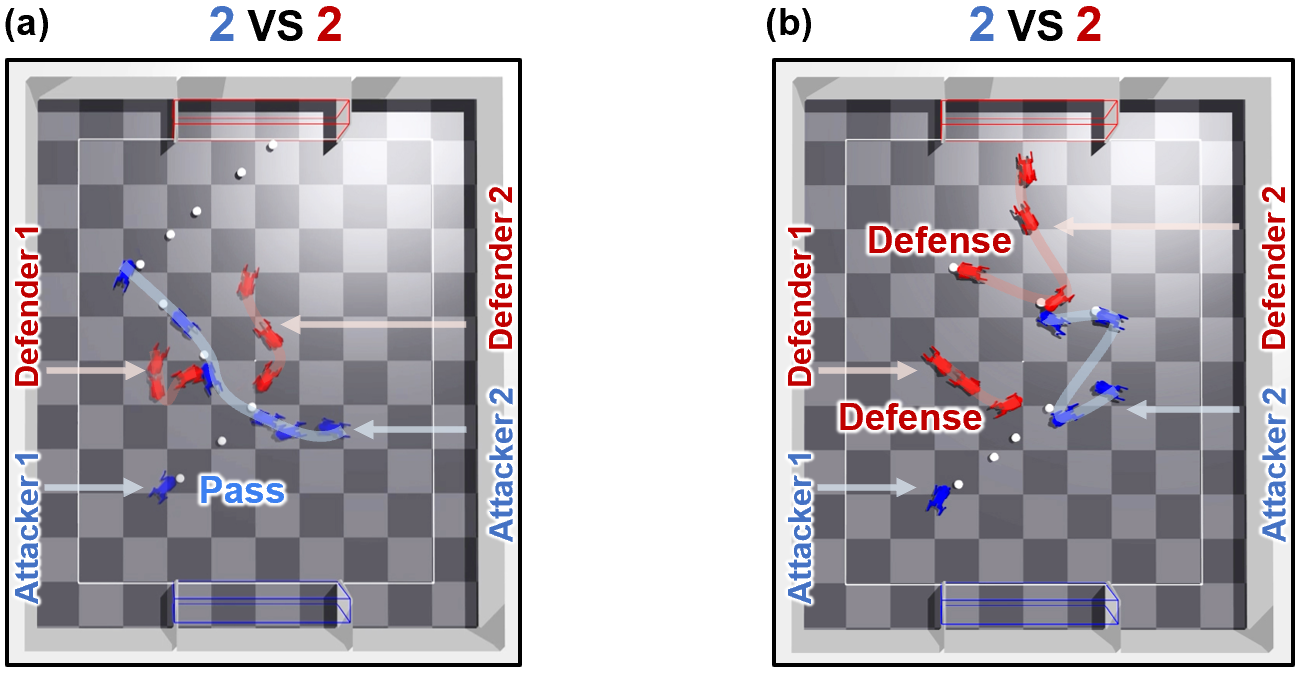}
    \caption{Demonstration of emergent strategies in 2v2 game. (a): The attackers executed coordinated attacks. (b) The defender performed effective coordinated defenses.}
    \label{fig:2v2}
\end{figure}

To evaluate the generalizability of our training pipeline, we extend the experiments to a 2v2 game in simulation, as shown in Fig.~\ref{fig:2v2}. Results show that our hierarchical framework combined with FSP scales effectively to more agents, yielding stable and coordinated attacking and defensive behaviors.

\section{Real-world Soccer System Details}
\label{app:system}
\begin{figure}[h]
	\vspace{0.0cm}
	\centering
	\begin{center}
		\includegraphics[width=0.9\columnwidth]{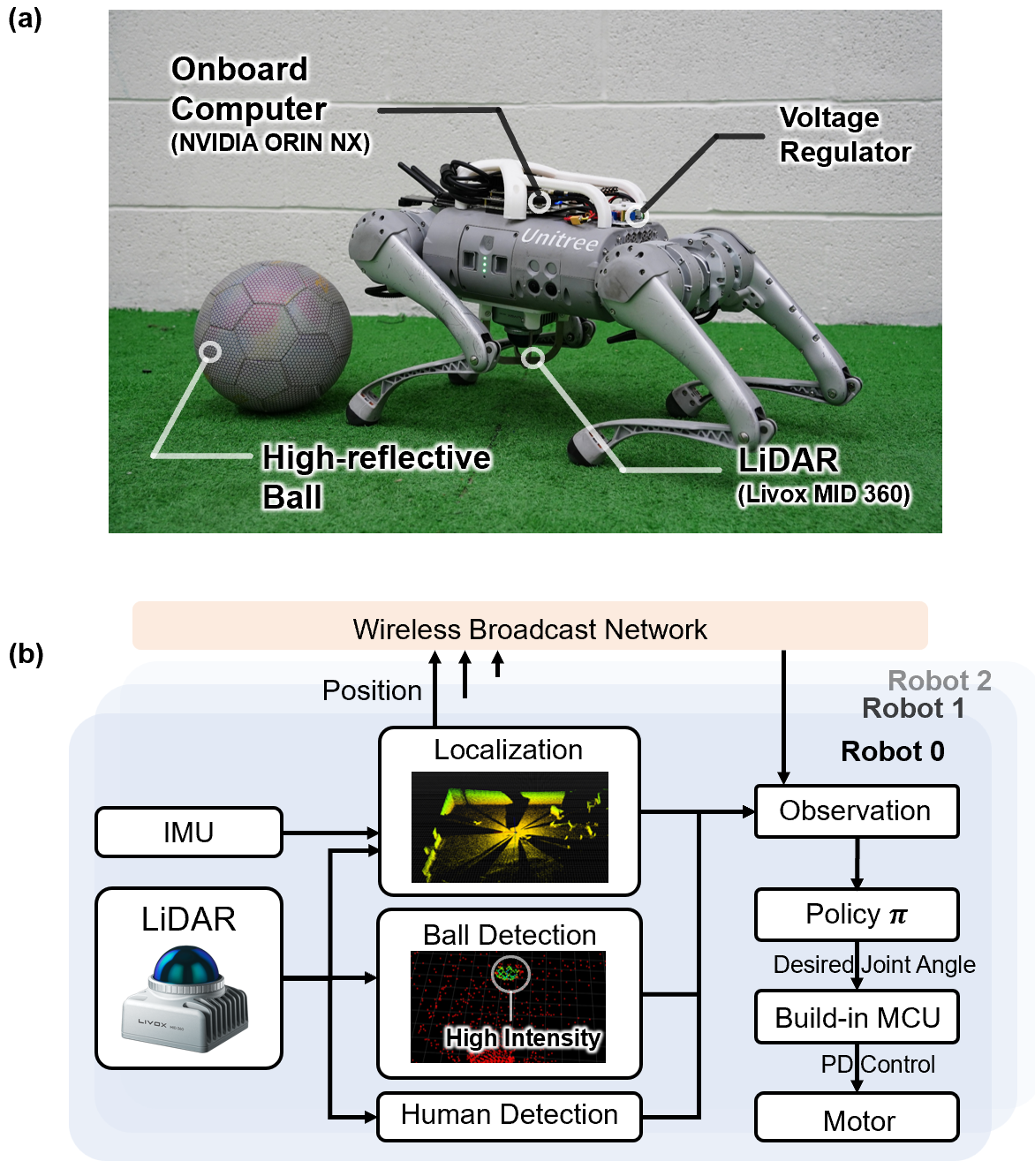}
	\end{center}
	\vspace{-0.3cm}
    \caption{Real world deployment overview: (a) \textbf{Quadrupedal Soccer Hardware}: Each robot is equipped with a LiDAR and onboard computer for perception and control. (b) \textbf{Software Pipeline}: The system performs ball detection, human detection, and self-localization from LiDAR and IMU data. This results in a decentralized architecture where each robot makes decisions independently based on local observations, with only lightweight position sharing to assist mutual localization.}
	\label{fig:real-world-system}
\end{figure}

We present the hardware and software pipeline used in our real-world quadruped soccer system. 
As shown in Figure~\ref{fig:real-world-system}(a), each Unitree Go1 robot is equipped with a Livox MID-360 LiDAR and interacts with a high-reflectivity ball. 
An onboard NVIDIA Orin NX computer performs real-time localization and detection computing and policy inference.
Both the LiDAR and Orin are powered via an integrated voltage regulator connected to the robot’s internal power supply.

Figure~\ref{fig:real-world-system}(b) depicts the software architecture. Both the indoor and outdoor self-localization are achieved using FAST-LIO~\citep{xu2021fast}, a computationally efficient and robust
inertial LiDAR odometry package, within a pre-mapping environment.
The ball is detected by filtering high-intensity LiDAR points resulting from its reflective surface, while human detection relies on filtering points based on their height.
The position of each robot is shared through a wireless broadcast network. 
Each robot constructs its own observation and feeds it into a learned policy $\boldsymbol\pi$, which integrates both high-level strategic decision-making and low-level motor skills, enabling decentralized soccer play. The resulting joint position commands are executed by the onboard microcontroller unit (MCU) using PD control to actuate the motors.

\end{document}